\documentclass[11pt,a4paper]{article}
\usepackage[hyperref]{naaclhlt2019}
\usepackage{times}
\usepackage{latexsym}
\usepackage{times}  %Required
\usepackage{helvet}  %Required
\usepackage{courier}  %Required
\usepackage{url}  %Required
\usepackage{graphicx}  %Required
\usepackage{multirow}
\usepackage{amsmath}
\usepackage{booktabs}
\usepackage{enumitem}
\usepackage{multicol}
\usepackage{subfigure}
\usepackage{caption}
\usepackage{float}
\usepackage{CJKutf8}

\usepackage{url}
\aclfinalcopy
\def\aclpaperid{839} %  Enter the acl Paper ID here

\title{Code-Switching for Enhancing NMT with Pre-Specified Translation}

\author{Kai Song\textsuperscript{1,2},
Yue Zhang\textsuperscript{3},
Heng Yu\textsuperscript{2},
Weihua Luo\textsuperscript{2},
Kun Wang\textsuperscript{1},
Min Zhang\textsuperscript{1}\\
\textsuperscript{1} Soochow University, Suzhou, China\\
\textsuperscript{2} Alibaba DAMO Academy, Hangzhou, China\\
\textsuperscript{3} School of Engineering, Westlake University, Hangzhou, China\\
{\tt songkai.sk@alibaba-inc.com}, {\tt zhangyue@wias.org.cn}\\
{\tt \{yuheng.yh,weihua.luowh\}@alibaba-inc.com}\\
{\tt kwang1994@stu.suda.edu.cn}, {\tt minzhang@suda.edu.cn}\\
}

\date{}

\begin{document}

\maketitle

\begin{abstract}
Leveraging user-provided translation to constrain NMT has practical significance. Existing methods can be classified into two main categories, namely the use of placeholder tags for lexicon words and the use of hard constraints during decoding. Both methods can hurt translation fidelity for various reasons. We investigate a data augmentation method, making code-switched training data by replacing source phrases with their target translations. Our method does not change the NMT model or decoding algorithm, allowing the model to learn lexicon translations by copying source-side target words. Extensive experiments show that our method achieves consistent improvements over existing approaches, improving translation of constrained words without hurting unconstrained words.
\end{abstract}

\section{Introduction}
\begin{figure}[t]
\center
\includegraphics[width=0.48\textwidth]{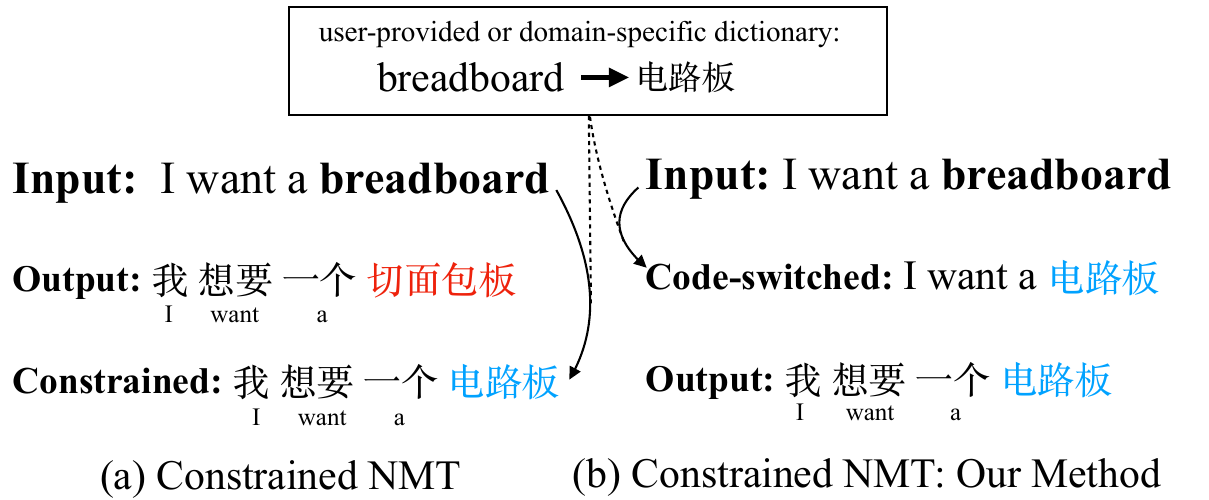}
\caption{Constrained NMT}
\label{fig:intro-fig}
\end{figure}

\begin{CJK*}{UTF8}{gbsn}
One important research question in domain-specific machine translation \cite{luong2015stanford} is how to impose translation constraints (\citeauthor{crego2016systran}, \citeyear{crego2016systran}; \citeauthor{hokamp2017lexically}, \citeyear{hokamp2017lexically}; \citeauthor{post2018fast}, \citeyear{post2018fast}). As shown in Figure \ref{fig:intro-fig} (a), the word ``breadboard'' can be translated into ``切面包板\ (a wooden board that is used to cut bread on)" in the food domain, but ``电路板\ (a construction base for prototyping of electronics)" in the electronic domain. To enhance translation quality, a lexicon can be leveraged for domain-specific or user-provided words (\citeauthor{arthur2016incorporating}, \citeyear{arthur2016incorporating}; \citeauthor{hasler2018neural}, \citeyear{hasler2018neural}). We investigate the method of leveraging pre-specified translation for NMT using such a lexicon.
\end{CJK*}

 For leveraging pre-specified translation, one existing approach uses {\it placeholder tags} to substitute named entities (\citeauthor{crego2016systran}, \citeyear{crego2016systran}; \citeauthor{li2016neural}, \citeyear{li2016neural}; \citeauthor{wang2017sogou}, \citeyear{wang2017sogou}) or rare words \cite{luong2014addressing} on both the source and target sides during training, so that a model can translate such words by learning to translate placeholder tags. For example, the $i$-th named entity in the source sentence is replaced with ``$tag_i$", as well as its corresponding translation in the target side. Placeholder tags in the output are replaced with pre-specified translation as a post-processing step. One disadvantage of this approach, however, is that the meaning of the original words in the pre-specified translation is not fully retained, which can be harmful to both adequacy and fluency of the output.

Another approach  (\citeauthor{hokamp2017lexically}, \citeyear{hokamp2017lexically}; \citeauthor{post2018fast}, \citeyear{post2018fast}) imposes pre-specified translation via {\it lexical constraints}, making sure such constraints are satisfied by modifying NMT decoding. This method ensures that pre-specified translations appear in the output. A problem of this method is that it does not explicitly explore the correlation between pre-specified translations and their corresponding source words during decoding, and thus can hurt translation fidelity \cite{hasler2018neural}. There is not a mechanism that allows the model to {\it learn} constraint translations during training, which the placeholder method allows.

We investigate a novel method based on {\it data augmentation}, which combines the advantages of both methods above. The idea is to construct synthetic parallel sentences from the original parallel training data. The synthetic sentence pairs resemble code-switched source sentences and their translations, where certain source words are replaced with their corresponding target translations. The motivation is to make the model learn to ``translate" embedded pre-specified translations by copying them from the modified source. During decoding, the source is similarly modified as a pre-processing step. As shown in Figure \ref{fig:intro-fig} (b), translation is executed over the code-switched source, without further constraints or post-processing.

In contrast to the placeholder method, our method keeps lexical semantic information (i.e. target words v.s. placeholder tags) in the source, which can lead to more adequate translations. Compared with the lexical constraint method, pre-specified translation is {\it learned} because such information is available both in training and decoding. As a data augmentation method, it can be used on any NMT architecture. In addition, our method enables the model to translate code-switched source sentences, and preserve its strength in translating un-replaced sentences.

To further strengthen copying, we propose two model-level adjustments: First, we share target-side embeddings with source-side target words, so that target vocabulary words have a unique embedding in the NMT system. Second, we integrate pointer network (\citeauthor{vinyals2015pointer}, \citeyear{vinyals2015pointer}; \citeauthor{gulcehre2016pointing}, \citeyear{gulcehre2016pointing}; \citeauthor{gu2016incorporating}, \citeyear{gu2016incorporating}; \citeauthor{see2017get}, \citeyear{see2017get}) into the decoder. The copy mechanism was firstly proposed to copy source words. In our method, it is further used to copy source-side {\it target} words.

Results on large scale English-to-Russian (En-Ru) and Chinese-to-English (Ch-En) tasks show that our method outperforms both placeholder and lexical constraint methods over a state-of-the-art Transformer \cite{DBLP:journals/corr/VaswaniSPUJGKP17} model on various test sets across different domains. We also show that shared embedding and pointer network can lead to more successful applications of the copying mechanism. We release four high-quality En-Ru e-commerce test sets translated by Russian language experts, totalling 7169 sentences with an average length of 21\footnote{To best of our knowledge, this is the first public e-commerce test set.}.

\section{Related Work}
\textbf{Using placeholders.} \citeauthor{luong2014addressing} (\citeyear{luong2014addressing}) use {\it annotated unk tags} to present the {\it unk} symbols in training corpora, where the correspondence between source and target {\it unk} symbols are obtained from word alignment \cite{brown1993mathematics}. Output {\it unk tags} are replaced through a post-processing stage by looking up a pre-specified dictionary or copying the corresponding source word. \citeauthor{crego2016systran} (\citeyear{crego2016systran}) extended {\it unk tags} symbol to specific symbols that can present name entities. \citeauthor{wang2017sogou} (\citeyear{wang2017sogou}) and \citeauthor{li2016neural} (\citeyear{li2016neural}) use a similar method. This method is limited when constrain NMT with pre-specified translations consisting of more general words, due to the loss of word meaning when representing them with placeholder tags. In contrast to their work, word meaning is fully kept in modified source in our work.

\textbf{Lexical constraints.} \citeauthor{hokamp2017lexically} (\citeyear{hokamp2017lexically}) propose an altered beam search algorithm, namely grid beam search, which takes target-side pre-specified translations as lexical constraints during beam search. A potential problem of this method is that translation fidelity is not specifically considered, since there is no indication of a matching source of each pre-specific translation. In addition, decoding speed is significantly reduced \cite{post2018fast}. \citeauthor{hasler2018neural} (\citeyear{hasler2018neural}) use alignment to gain target-side constraints' corresponding source words, simultaneously use finite-state machines and multi-stack \cite{DBLP:journals/corr/AndersonFJG16a} decoding to guide beam search. \citeauthor{post2018fast} (\citeyear{post2018fast}) give a fast version of \citeauthor{hokamp2017lexically} (\citeyear{hokamp2017lexically}), which limits the decoding complexity linearly by altering the beam search algorithm through dynamic beam allocation.

In contrast to their methods, our method does not make changes to the decoder, and therefore decoding speed remains unchanged. Translation fidelity of pre-specified source words is achieved through a combination of training and decoding procedure, where replaced source-side words still contain their target-side meaning. As a soft method of inserting pre-specified translation, our method does not guarantee that all lexical constraints are satisfied during decoding, but has better overall translation quality compared to their method.

\textbf{Using probabilistic lexicons.} Aiming at making use of one-to-many phrasal translations, the following work is remotely related to our work. \citeauthor{tang2016neural} (\citeyear{tang2016neural}) use a phrase memory to provide extra information for their NMT encoder, dynamically switching between word generation and phrase generation during decoding. \citeauthor{wang2017neural} (\citeyear{wang2017neural}) use SMT to recommend prediction for NMT, which contains not only translation operations of a SMT phrase table, but also alignment information and coverage information. \citeauthor{arthur2016incorporating} (\citeyear{arthur2016incorporating}) incorporate discrete lexicons by converting lexicon probabilities into predictive probabilities and linearly interpolating them with NMT probability distributions. 

Our method is similar in the sense that external translations of source phrases are leveraged. However, their tasks are different. In particular, these methods regard one-to-many translation lexicons as a {\it suggestion}. In contrast, our task aims to {\it constrain} NMT translation through \textbf{one-to-one} pre-specified translations. Lexical translations can be used to generate code-switched source sentences during training, but we do not modify NMT models by integrating translation lexicons. In addition, our data augmentation method is more flexible, because it is model-free.

\citeauthor{alkhouli2018alignment} (\citeyear{alkhouli2018alignment}) simulate a dictionary-guided translation task to evaluate NMT's alignment extraction. A one-to-one word translation dictionary is used to guide NMT decoding. In their method, a dictionary entry is limited to only one word on both the source and target sides. In addition, a pre-specified translation can come into effect only if the corresponding source-side word is successfully aligned during decoding.

On translating named entities, \citeauthor{currey2017copied} (\citeyear{currey2017copied}) augment the training data by copying target-side sentences to the source-side, resulting in augmented training corpora where the source and the target sides contain identical sentences. The augmented data is shown to improve translation performance, especially for proper nouns and other words that are identical in the source and target languages.

\section{Data augmentation}  \label{augmentation}
Our method is based on {\it data augmentation}. During training, augmented data are generated by replacing source words or phrases directly with their corresponding target translations. The motivation is to sample as many code-switched translation pairs as possible. During decoding, given pre-specified translations, the source sentence is modified by replacing phrases with their pre-specified translations, so that the trained model can directly copy embedded target translations in the output.

\subsection{Training} \label{training}
Given a bilingual training corpus, we sample augmented sentence pairs by leveraging a SMT phrase table, which can be trained over the same bilingual corpus or a different large corpus. We extract source-target phrase pairs\footnote{Source-side phrase is at most trigram.} from the phrase table, replacing source-side phrases of source sentences using the following sampling steps:
\begin{enumerate}
\item Indexing between source-target phrase pairs and training sentences: (a) For each source-target phrase pair, we record all the matching bilingual sentences that contain both the source and target. Word alignment can be used to ensure the phrase pairs that are mutual translation. (b) We also sample bilingual sentences that match two source-target phrase pairs. In particular, given a combination of two phrase pairs, we index bilingual sentences that match both simultaneously.
\item Sampling:
(a) For each source-target phrase pair, we keep at most $k_1$ randomly selected matching sentences. The source-side phrase is replaced with its target-side translation. (b) For each combination of two source-target phrase pairs, we randomly sample at most $k_2$ matching sentences. Both source-side matching phrases are replaced with their target translations.\footnote{We set $k_1=100, k_2=30$ empirically.}
\end{enumerate}

The sampled training data is added to the original training data to form a final set of training sentences.

\subsection{Decoding} \label{decoding}
We impose target-side pre-specified translations to the source by replacing source phrases with their translations. Lexicons are defined in the form of one-to-one source-target phrase pairs. Different from training, the number of replaced phrases in a source sentence is not necessarily restricted to one or two, which will be discussed in Section \ref{analysis}. In practice, pre-specified translations can be provided by customers or through user feedback, which contains one identified translation for specified source segment.

\section{Model}
\textbf{Transformer} \cite{DBLP:journals/corr/VaswaniSPUJGKP17} uses self-attention network for both encoding and decoding. The encoder is composed of $n$ stacked neural layers. For time step $i$ in layer $j$, the hidden state $h_{i,j}$ is calculated by employing self-attention over the hidden states in layer $j-1$, which are $\{h_{1,j-1}, h_{2,j-1}, ..., h_{m,j-1}\}$, where $m$ is the number of source-side words.

In particular, $h_{i,j}$ is calculated as follows: First, a self-attention sub-layer is employed to encode the context. Then {\it attention weights} are computed as scaled dot product between the current query  $h_{i,j-1}$ and all keys $\{h_{1,j-1}, h_{2,j-1}, ..., h_{m,j-1}\}$, normalized with a {\rm softmax} function. After that, the context vector is represented as {\it weighted sum} of the values projected from hidden states in the previous layer, which are $\{h_{1,j-1}, h_{2,j-1}, ..., h_{m,j-1}\}$. The hidden state in the previous layer and the context vector are then connected by residual connection, followed by a layer normalization function \cite{ba2016layer}, to produce a candidate hidden state $h_{i,j}^{'}$. Finally, another sub-layer including a feed-forward network (FFN) layer, followed by another residual connection and layer normalization, are used to obtain the hidden state $h_{i,j}$.

In consideration of translation quality, multi-head attention is used instead of single-head attention as mentioned above, positional encoding is also used to compensate the missing of position information in this model.

The decoder is also composed of $n$ stacked layers. For time step $t$ in layer $j$, a self-attention sub-layer of hidden state $s_{t,j}$ is calculated by employing self-attention mechanism over hidden states in previous target layer, which are  $\{s_{1,j-1}, s_{2,j-1}, ..., s_{t-1,j-1}\}$, resulting in candidate hidden state $s_{t,j}^{'}$. Then, a second target-to-source sub-layer of hidden state $s_{t,j}$ is inserted above the target self-attention sub-layer. In particular, the queries($Q$) are projected from $s_{t,j}^{'}$, and the keys($K$) and values($V$) are projected from the source hidden states in the last layer of encoder, which are $\{h_{1,n}, h_{2,n}, ..., h_{m,n}\}$.  The output state is another candidate hidden state $s_{t,j}^{''}$. Finally, a last feed-forward sub-layer of hidden state $s_{t,j}$ is calculated by employing self-attention over $s_{t,j}^{''}$.

 A {\rm softmax} layer based on decoder's last layer $s_{t,n}$ is used to gain a probability distribution $P_{\it predict}$ over target-side vocabulary.
\begin{equation}
\label{softmax}
 \begin{split}
 p(y_t|&y_{1},...,y_{t-1},\textbf{x}) = {\rm softmax}(s_{t,n} * \textbf{W}),
 \end{split}
\end{equation}
where $\textbf{W}$ is the weight matrix which is learned, $\textbf{x}$ represent the source sentence, $\{y_{1}, y_{2}, ..., y_{t}\}$ represent target words.

\begin{figure}[t]
\center
\includegraphics[width=0.49\textwidth]{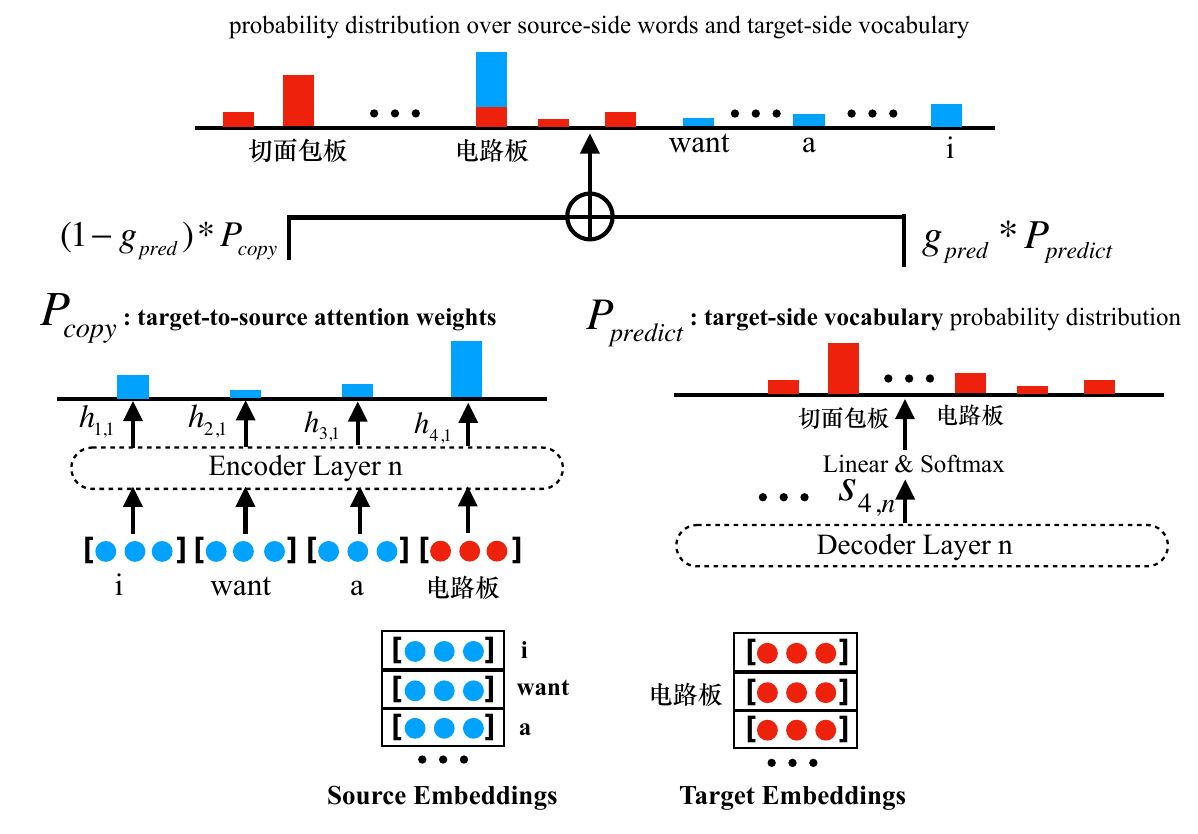}
\caption{Shared embeddings and pointer network}
\label{fig:pointing}
\end{figure}

\subsection{Shared Target Embeddings} 
Shared target embeddings enforces the correspondence between source-side and target-side expressions on the embedding level. As shown in Figure \ref{fig:pointing}, during encoding, source-side target word embeddings are identical to their embeddings in the target-side vocabulary embedding matrix. This makes it easier for the model to copy source-side target words to the output.
%To our knowledge, we are the first to investigate mixing source and target embeddings in code-switched language encoding.

\subsection{Pointer Network}
To strengthen copying through locating source-side target words, we integrate {\it pointer network} \cite{gulcehre2016pointing} into the decoder, as shown in Figure \ref{fig:pointing}. At each decoding time step $t$, the target-to-source attention weights $\alpha_{t,1}, ..., \alpha_{t,m}$ are utilized as a probability distribution $P_{copy}$, which models the probability of {\it copying} a word from the $i$-th source-side position. The $i$-th source-side position may represent a source-side word or a source-side target word. $P_{copy}$ is added to $P_{predict}$, the probability distribution over target-side vocabulary, to gain a new distribution over both the source and the target side vocabulary\footnote{For the words which belong to the source-side vocabulary but are not appeared in the source-side sentence, the probabilities are set to 0.}:
\begin{equation}
\label{objective}
    P=(1-g_{pred})*P_{\it copy} + g_{pred}*P_{\it predict},
\end{equation}
where $g_{pred}$ is used to control the contribution of two probability distributions. For time step $t$, $g_{pred}$ is calculated from the context vector $c_t$ and the current hidden state of the decoder's last layer $s_{t,n}$:
\begin{equation}
\label{copy_predict}
    g_{pred} = \sigma (c_t * W_p + s_{t,n} * W_q + b_r),
\end{equation}
where $W_p$, $W_q$, and  $b_r$ are parameters trained and $\sigma$ is the sigmoid function. In addition, the context vector $c_t$ is calculated as $ c_t = \sum_{i=1}^m \alpha_{t,i}*h_{i,n} $, where $\alpha_{t,i}$ is attention weight mentioned earlier. $\{h_{1,n}, h_{2,n}, ..., h_{m,n}\}$ are the source-side hidden states of the encoder's last layer.

\section{Experiments} \label{experiments}
We compare our method with strong baselines on large-scale En-Ru and Ch-En tasks on various test sets across different domains, using a strongly optimized Transformer \cite{DBLP:journals/corr/VaswaniSPUJGKP17}. BLEU \cite{papineni:2002} is used for evaluation.

\subsection{Data}
Our training corpora are taken from the WMT2018 news translation task.

\textbf{En-Ru.} We use 13.88M sentences as baseline training data, containing both a real bilingual corpus and a synthetic back-translation corpus \cite{Sennrich2015Improving}. The synthetic corpus is translated from ``NewsCommonCrawl", which can be obtained from the WMT task. The news domain contains four different test sets published by WMT2018 over the recent years, namely ``news2015'', ``news2016", ``news2017", and ``news2018", respectively, each having one reference. The e-commerce domain contains four files totalling 7169 sentences, namely ``subject17'', ``desc17", ``subject18", and ``desc18", respectively, each having one reference. The sentences are extracted from e-commerce websites, in which ``subject"s are the goods names shown on a listing page. ``desc"s refer to information in a commodity's description page. ``subject17'' and ``desc17" are released\footnote{\url{https://github.com/batman2013/e-commerce_test_sets}}. Our development set is ``news2015''.

\textbf{Ch-En.} We use 7.42M sentences as our baseline training data, containing both real bilingual corpus and synthetic back-translation corpus \cite{Sennrich2015Improving}. We use seven public development and test data sets, four in the news domain, namely ``NIST02", ``NIST03", ``NIST04", ``NIST05", respectively, each with four references, and three in the spoken language domain, namely ``CSTAR03", ``IWSLT2004", ``IWLST2005", respectively, each with 16 references. ``NIST03" is used for development.

\subsection{Experimental Settings} \label{experimentalsettings}
We use six self-attention layers for both the encoder and the decoder. The embedding size and the hidden size are set to 512. Eight heads are used for self-attention. A feed-forward layer with 2048 cells and Swish \cite{ramachandran2018searching} is used as the activation function. Adam \cite{kingma2014adam} is used for training; warmup step is 16000; the learning rate is 0.0003. We use label smoothing \cite{junczys2016amu} with a confidence score of 0.9, and all the drop-out \cite{gal2016theoretically} probabilities are set to 0.1.

We extract a SMT phrase table on the bilingual training corpus by using moses~\cite{koehn2007moses} with default setting, which is used for matching sentence pairs to generate augmented training data. We apply count-based pruning \cite{zens2012systematic} to the phrase table, the threshold is set to 10.

During \textbf{decoding}, similar to \citeauthor{hasler2018neural} (\citeyear{hasler2018neural}), \citeauthor{alkhouli2018alignment} (\citeyear{alkhouli2018alignment})  and \citeauthor{post2018fast} (\citeyear{post2018fast}), we make use of references to obtain gold constraints. Following previous work, pre-specified translations for each source sentence are sampled from references and used by all systems for fair comparison.

In all the baseline systems, the vocabulary size is set to 50K on both sides. For ``Data augmentation", to allow the source-side dictionary to cover target-side words, the target- and source-side vocabularies are merged for a new source vocabulary. For ``Shared embeddings", the source vocabulary remains the same as the baselines, where the source-side target words use embeddings from target-side vocabulary.
\begin{table*}[t]
 \centering\setlength{\tabcolsep}{2.0 pt}
    \begin{tabular}{l|ccccc|ccccc}
    \toprule
    &news15 & news16 & news17&  news18&$\triangle{}$ & subject17 & desc17 & subject18 & desc18& $\triangle{}$ \\
    \midrule
    Marian & 33.27 &  31.91 & 36.18 & 32.11 & -0.15 & 8.03  & 23.21 & 11.02 &  27.94 & -0.46\\ 
     \midrule
    Transformer &33.29&  31.95&  36.57&  32.27& -&  8.56&  23.53&  11.95&  27.90 & -\\ 
    + Placeholder & 33.14&  32.07&  36.24&  32.03&  -0.15& 9.81&  24.04&  13.84&  29.34 &+1.27\\
    + Lexi. Cons. &  33.50&  32.62&  36.65&  32.88&  +0.39&9.24&  23.67&  13.1&  29.83& +0.98\\
    \midrule
    Data Aug. & 34.71&  33.69&  38.43&  33.51& +1.57& 10.63&  25.56&  14.26&  30.92& +2.36\\
    + Share & 35.28&  34.37&  39.02&  34.44& +2.26& 10.82&  25.84&  15.20&  30.97& +2.72\\
    + Share\&Point & \textbf{36.44}&  \textbf{35.31}&  \textbf{40.23}&  \textbf{35.43}&\textbf{+3.33}&  \textbf{11.58}&  \textbf{26.53}&  \textbf{16.08}& \textbf{32.17}& \textbf{+3.61}\\
\bottomrule
\end{tabular}
\caption{Results on En-Ru, one or two source phrases of each sentence have pre-specified translation. ``Transformer" is our in-house vanilla Transformer baseline. ``Marian" is the implementation of Transformer by \citeauthor{mariannmt} (\citeyear{mariannmt}), which is used as a reference of our Transformer implementation.}
\label{tab:enru_results}
\end{table*}

\begin{table*}[]
    \centering\setlength{\tabcolsep}{2.8 pt}
    \begin{tabular}{l|cccc|ccccc}
         \toprule
         & CSTAR03 &IWSLT04 & IWSLT05 &$\triangle{}$&NIST02 &NIST03 &NIST04 &NIST05& $\triangle{}$ \\
          \midrule
         Transformer & 53.03&  56.52&  64.72& -& 40.52&  37.85&  40.12&  39.26& -\\
         + Placeholder &52.51&  56.15&  64.44& -0.39&  40.01&  37.16&  39.96&  38.87 & -0.44\\
         + Lexi. Cons. &53.30&  56.95&  65.63& +0.54&  40.36&  38.02&  40.44&  39.72 & +0.20\\
         \midrule
        Data Aug. &53.82&  57.28&  65.54& +0.79&  40.85&  38.41&  40.81&  40.29& +0.65\\
        +Share &\textbf{53.90}&  \textbf{57.67}&  65.59& \textbf{+0.96}& 41.06&  38.57&  41.22&  40.38& +0.87\\
        +Share\&Point &53.79&  57.29&  \textbf{65.65}& +0.82& \textbf{41.11}&  \textbf{38.7}&  \textbf{41.3}&  \textbf{40.4}& \textbf{+0.94}\\
        \bottomrule
    \end{tabular}
    \caption{Results on Ch-En, one or two source phrases of each sentence have pre-specified translation.}
    \label{tab:zhen_results}
\end{table*}

\subsection{System Configurations}
We use an in-house reimplementation of Transformer, similar to Google's Tensor2Tensor. For the baselines, we reimplement \citeauthor{crego2016systran} (\citeyear{crego2016systran}), as well as \citeauthor{post2018fast} (\citeyear{post2018fast}). BPE \cite{sennrich2015neural} is used for all experiments, the operation is set to 50K. Our test sets cover news and e-commerce domains on En-Ru, and news and spoken language domains on Ch-En.

\textbf{Baseline 1: Using Placeholder.} We combine \citeauthor{luong2014addressing} (\citeyear{luong2014addressing}) and \citeauthor{crego2016systran} (\citeyear{crego2016systran}). For generating placeholder tags during training, following \citeauthor{crego2016systran} (\citeyear{crego2016systran}), we use a named entity translation dictionary which is extracted from Wikidata\footnote{\url{https://www.wikidata.org}}. The dictionary is released together with e-commerce test sets, which is mentioned before. For Ch-En, the dictionary contains 285K person names, 746K location names and 1.6K organization names. For En-Ru, the dictionary contains 471K person names, 254K location names and 1.5K organization names. Additionally, we manually corrected  a dictionary which contains 142K brand names and product names translation for En-Ru. By further leveraging word alignment in the same way as \citeauthor{luong2014addressing} (\citeyear{luong2014addressing}), the placeholder tags are annotated with indices. We use FastAlign~\cite{dyer2013simple} to generate word alignment. The amount of sentences containing placeholder tags is controlled to a ratio of 5\% of the corpus. During decoding, pre-specified translations described in Section \ref{experimentalsettings} are used.

\textbf{Baseline 2: Lexical Constraints.} We re-implement \citeauthor{post2018fast} (\citeyear{post2018fast}), integrating their algorithm into our Transformer. Target-side words or phrases of pre-specified translations mentioned in Section \ref{experimentalsettings} are used as lexical constraints.

\textbf{Our System.} During training, we use the method described in Section \ref{training} to obtain the augmented training data. The SMT phrase table mentioned in Section \ref{experimentalsettings} is used for ``Indexing" and ``Sampling". During decoding, pre-specified translations mentioned in Section \ref{experimentalsettings} are used. The augmented data contain sampled sentences with one or two replacements on the source side. By applying the two sampling steps described in Section \ref{training}, about 10M and 6M augmented Ch-En and En-Ru sentences are generated, respectively. The final training corpora consists of both the augmented training data and the original training data.
\begin{figure*}[t]
\center
\includegraphics[width=1\textwidth]{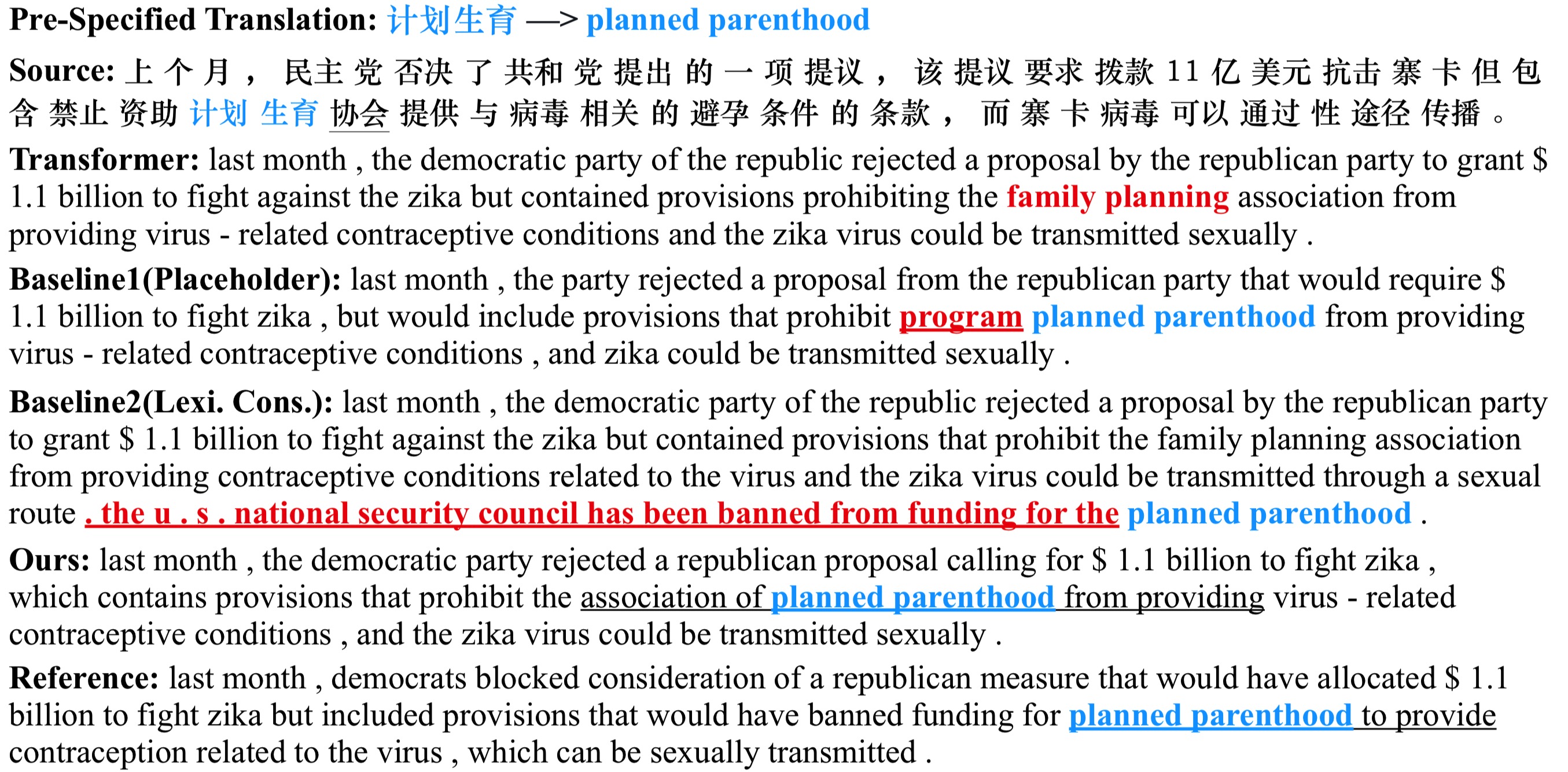}
\caption{Sample outputs.}
\label{fig:sample-fig1}
\end{figure*}

\subsection{Results} \label{results}
\textbf{Comparison with Baselines.} Our Transformer implementation can give comparable performance with state-of-the-art NMT~\cite{mariannmt}, see ``Transformer" and ``Marian" in Table \ref{tab:enru_results}, which also shows a comparison of different methods on En-Ru. The {\it lexical constraint} method gives improvements on both the news and the e-commerce domains, compared with the Transformer baseline. The {\it placeholder} method also gives an improvement on the e-commerce domain. The average improvement is calculated over all the test set results in each domain. In the news domain, the average improvement of {\it our method} is 3.48 BLEU higher compared with {\it placeholder}, and 2.94 over {\it lexical constraints}. In the e-commerce domain, the average improvement of {\it our method} is 1.34 BLEU compared with {\it placeholder}, and 2.63 with {\it lexical constraints}. Both shared embedding and pointer network are effective. Table \ref{tab:zhen_results} shows the same comparison on Ch-En. In the spoken language domain, the average improvement is 1.35 BLEU compared with {\it placeholder}, and 0.42 with {\it lexical constraints}. In the news domain, the average improvement is 1.38 BLEU compared with {\it placeholder}, and 0.74 with {\it lexical constraints}.

We find that the {\it placeholder} method can only bring improvements on the En-Ru e-commerce test sets, since the pre-specified translations of the four e-commerce test sets are mostly {\it entities}, such as brand names or product names. Using placeholder tags to represent these entities leads to relatively little loss of word meaning. But on many of the other test sets, pre-specified translations are mostly vocabulary words. The placeholder tags fail to keep their word meaning during translation, leading to lower results.
\begin{table}[t]
    \centering\setlength{\tabcolsep}{4 pt}
    \begin{tabular}{c|cccc}
    \toprule
        Beam Size & 5& 10 & 20 &  30\\
         \midrule
         Unconstrained \& Ours &  416 &  312 &  199 &  146 \\
         Lexical Constraint & 102 &   108 &  74 &  50 \\
         \bottomrule
    \end{tabular}
    \caption{Decoding speed (words/sec), Ch-En dev set.}
    \label{tab:speed}
\end{table}

The speed contrast between unconstrained NMT, lexical constraint and our method is shown in Table \ref{tab:speed}. The decoding speed of our method is equal to unconstrained NMT, and faster than the lexical constraint method, which confirms our intuition introduced earlier.

\begin{CJK*}{UTF8}{gbsn}
\textbf{Sample Outputs.} Figure \ref{fig:sample-fig1} gives a comparison of different system's translations. Given a Chinese source sentence, the baseline system fails to translate ``计划生育" adequately, as ``family planning" is not a correct translation of ``计划生育". In the pre-specified methods, the correct translation (``计划生育" to ``planned parenthood") is achieved through different ways.

For the {\it placeholder} method, the source phrase ``计划 生育" is replaced with the placeholder tag ``$tag_1$" during pre-processing. After translation, output ``$tag_1$" is replaced with ``planned parenthood" as a post-processing step. However, the underlined word ``program" is generated before ``planned parenthood", which has no relationship with any source-side word. The source-side word ``协会", which means ``association", is omitted in translation. Through deeper analysis, the specific phrase ``program $tag_1$" occurs frequently in the training data. During decoding, using the hard tag leads to the loss of the source phrase's original meaning. As a result, the word ``program" is incorrectly generated along with ``$tag_1$".

The {\it lexical constraints} method regards the target side of the pre-specified translation as a lexical constraint. Here the altered beam search algorithm fails to predict the constraint ``planned parenthood" during previous decoding steps. Although the constraint finally comes into effect, over translation occurs, which is highlighted by the underlined words. This is because the method enforces hard constraints, preventing decoding to stop until all constraints are met.

Our method makes use of pre-specified translation by replacing the source-side phrase ``计划 生育" with the target-side translation ``planned parenthood", copying the desired phrase to the output along with the decoding procedure. The translation ``association of planned parenthood from providing" is the exact translation of the source-side phrase ``计划(planned) 生育(parenthood) 协会(association) 提供(providing)", and agrees with the reference, ``planned parenthood to provide".
\end{CJK*}
\begin{figure}[t]
\centering\setlength{\tabcolsep}{4 pt}
\includegraphics[width=0.4\textwidth]{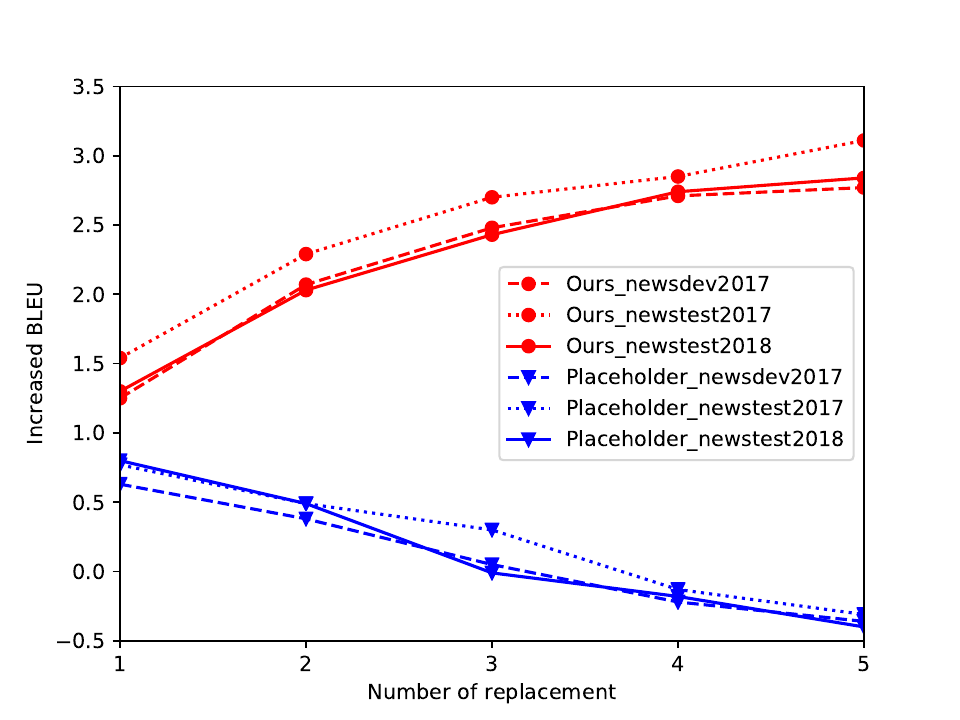}
\caption{Increased BLEU on Ch-En test sets.}
\label{fig:grow-bleu}
\end{figure}

\subsection{Analysis} \label{analysis}
 \begin{figure}[t]
\centering\setlength{\tabcolsep}{4 pt}
\includegraphics[width=0.4\textwidth]{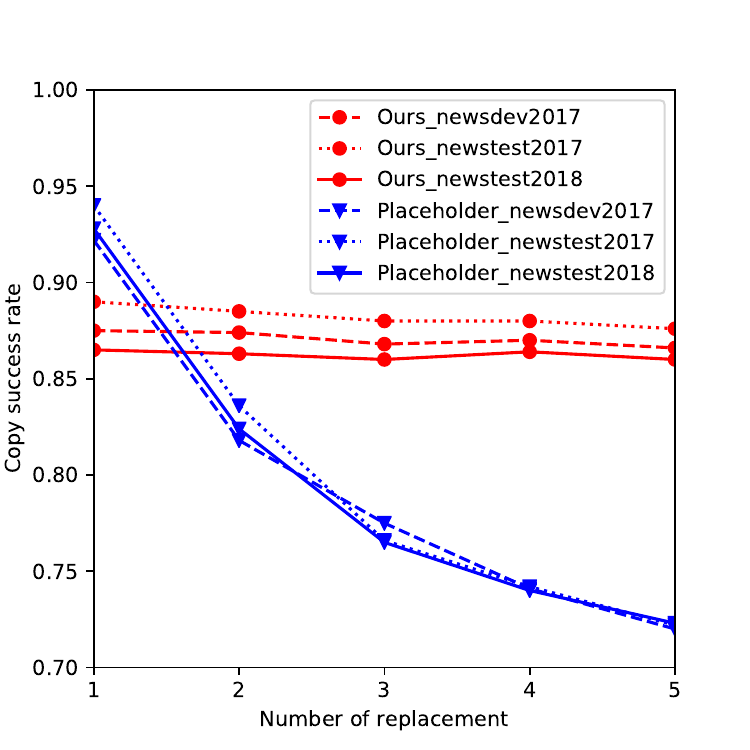}
\caption{Copy success rate on Ch-En test sets.}
\label{fig:grow-csr}
\end{figure}

\textbf{Effect of Using More Pre-specified Translations.} Even though the augmented training data have only one or two replacements on the source side, the model can translate a source sentence with up to five replacements. Figure \ref{fig:grow-bleu} shows that compared with unconstrained Transformer, the translation quality of our method keeps increasing when the number of replacements increases, since more pre-specified translations are used.

We additionally measure the effect on the Ch-En WMT test sets, namely ``newsdev2017", ``newstest2017", ``newstest2018", respectively, each having only one reference instead of four. The baseline BLEU scores on these three test sets are 18.49, 20.01 and 19.05, respectively. Our method gives BLEU scores of 20.56, 22.3, 21.08, respectively, when using one or two pre-specified translations for each sentence. The increased BLEU when utilizing different number of pre-specified translations is shown in Figure \ref{fig:grow-bleu}. We found that the improvements on WMT test sets are more significant than on NIST, since pre-specified translations are sampled from one reference only, enforcing the output to match this reference. The placeholder method does not give consistent improvements on news test sets, due to the same reason as mentioned earlier.

As shown in Figure \ref{fig:grow-csr}, the copy success rate of our method does not decrease significantly when the number of replacements grows. Here, a {\it copy success} refers a pre-specified target translation that can occur in the output. The placeholder method achieves a higher copy success rate than ours when the number of replacements is 1, but the copy success rate decreases when using more pre-specified translations. The copy success rate of the lexical constraint method is always 100\%, since it imposes hard constraints rather than soft constraints. However, as discussed earlier, overall translation quality can be harmed as a cost of satisfying decoding constraints by their method.

In the presented experiment results, the highest copy success rate of our method is 90.54\%, which means a number of source-side target words or phrases are not successfully copied to the translation output. This may be caused by the lack of training samples for certain target-side words or phrases. In En-Ru, we additionally train a model with augmented data that is obtained by matching an SMT phrase table without any pruning strategy. The copy success rate can reach 98\%, even without using ``shared embedding" and ``pointer network" methods. 

\textbf{Effect of Shared Embeddings and Pointer Network.} The gains of {\it shared embeddings} and {\it pointer network} are reflected in both the copy success rate and translation quality. As shown in Table \ref{tab:zhen_nist_rate}, when using one pre-specified translation for each source sentence, the copy success rate improves on various test sets by integrating shared embeddings and pointer network, demonstrating that more pre-specified translations come into effect. Table \ref{tab:enru_results} and Table \ref{tab:zhen_results} earlier show the improvement of translation quality.
\begin{table}[t]
    \centering\setlength{\tabcolsep}{1.5 pt}
    \begin{tabular}{c|cccc}
        \toprule
        & NIST02 & NIST03 &  NIST04 & NIST05\\
         \midrule
         Data Aug. &  83.89\% &  85.71\% &  86.71\% &  87.45\%\\
         +Share\&Point & 87.72\% & 88.31\% &  89.18\% &  90.54\% \\
         \bottomrule
    \end{tabular}
    \caption{Copy success rate on Ch-En test sets.}
    \label{tab:zhen_nist_rate}
\end{table}

\begin{table}[t]
    \centering\setlength{\tabcolsep}{4 pt}
    \begin{tabular}{c|cccc}
    \toprule
         & news15& news16 & news17&  news18\\
         \midrule
         Baseline&  33.29&  31.95&  36.57&  32.27 \\
         Ours & 33.53&  32.29&  36.54&  32.47\\
         \bottomrule
    \end{tabular}
    \caption{BLEU scores of non code-switched (original) input on En-Ru test sets.}
    \label{tab:enru_wmt_raw}
\end{table}

\textbf{Translating non Code-Switched Sentences.} Our method preserves its strength on translating non code-switched sentences. As shown in Table \ref{tab:enru_wmt_raw}, the model trained on the augmented corpus has comparable strength on translating un-replaced sentences as the model trained on the original corpus. In addition, on some test sets, our method is slightly better than the baseline when translating non code-switched source sentences. This can be explained from two aspects: First, the augmented data make the model more robust to perturbed inputs; Second, the pointer network makes the model better by copying certain source-side words \cite{gulcehre2016pointing}, such as non-transliterated named entities.

\section{Conclusion}
We investigated a data augmentation method for constraining NMT with pre-specified translations, utilizing code-switched source sentences and their translations as augmented training data. Our method allows the model to learn to translate source-side target phrases by ``copying" them to the output, achieving consistent improvements over previous lexical constraint methods on large NMT test sets. To the best of our knowledge, we are the first to leverage code switching for NMT with pre-specified translations.

\section{Future Work}
In the future, we will study how the copy success rate and the BLEU scores interact when different sampling strategies are taken to obtain augmented training corpus and when the amount of augmented data grows. Another direction is to validate the performance when applying this approach to language pairs that contain a number of identical letters in their alphabets, such as English to French and English to Italian.

\section*{Acknowledgments}
We thank the anonymous reviewers for their detailed and constructed comments. Yue Zhang is the corresponding author. The research work is supported by the National Natural Science Foundation of China (61525205). Thanks for Shaohui Kuang, Qian Cao, Zhongqiang Huang and Fei Huang for their useful discussion.
\bibliographystyle{acl_natbib}
\bibliography{naaclhlt2019}

\begin{thebibliography}{32}
\expandafter\ifx\csname natexlab\endcsname\relax\def\natexlab#1{#1}\fi

\bibitem[{Alkhouli et~al.(2018)Alkhouli, Bretschner, and
  Ney}]{alkhouli2018alignment}
Tamer Alkhouli, Gabriel Bretschner, and Hermann Ney. 2018.
\newblock On the alignment problem in multi-head attention-based neural machine
  translation.
\newblock \emph{arXiv preprint arXiv:1809.03985}.

\bibitem[{Anderson et~al.(2016)Anderson, Fernando, Johnson, and
  Gould}]{DBLP:journals/corr/AndersonFJG16a}
Peter Anderson, Basura Fernando, Mark Johnson, and Stephen Gould. 2016.
\newblock \href {http://arxiv.org/abs/1612.00576} {Guided open vocabulary image
  captioning with constrained beam search}.
\newblock \emph{CoRR}, abs/1612.00576.

\bibitem[{Arthur et~al.(2016)Arthur, Neubig, and
  Nakamura}]{arthur2016incorporating}
Philip Arthur, Graham Neubig, and Satoshi Nakamura. 2016.
\newblock Incorporating discrete translation lexicons into neural machine
  translation.
\newblock \emph{arXiv preprint arXiv:1606.02006}.

\bibitem[{Ba et~al.(2016)Ba, Kiros, and Hinton}]{ba2016layer}
Jimmy~Lei Ba, Jamie~Ryan Kiros, and Geoffrey~E Hinton. 2016.
\newblock Layer normalization.
\newblock \emph{arXiv preprint arXiv:1607.06450}.

\bibitem[{Brown et~al.(1993)Brown, Pietra, Pietra, and
  Mercer}]{brown1993mathematics}
Peter~F Brown, Vincent J~Della Pietra, Stephen A~Della Pietra, and Robert~L
  Mercer. 1993.
\newblock The mathematics of statistical machine translation: Parameter
  estimation.
\newblock \emph{Computational linguistics}, 19(2):263--311.

\bibitem[{Crego et~al.(2016)Crego, Kim, Klein, Rebollo, Yang, Senellart,
  Akhanov, Brunelle, Coquard, Deng et~al.}]{crego2016systran}
Josep Crego, Jungi Kim, Guillaume Klein, Anabel Rebollo, Kathy Yang, Jean
  Senellart, Egor Akhanov, Patrice Brunelle, Aurelien Coquard, Yongchao Deng,
  et~al. 2016.
\newblock Systran's pure neural machine translation systems.
\newblock \emph{arXiv preprint arXiv:1610.05540}.

\bibitem[{Currey et~al.(2017)Currey, Barone, and Heafield}]{currey2017copied}
Anna Currey, Antonio Valerio~Miceli Barone, and Kenneth Heafield. 2017.
\newblock Copied monolingual data improves low-resource neural machine
  translation.
\newblock In \emph{Proceedings of the Second Conference on Machine
  Translation}, pages 148--156.

\bibitem[{Dyer et~al.(2013)Dyer, Chahuneau, and Smith}]{dyer2013simple}
Chris Dyer, Victor Chahuneau, and Noah~A Smith. 2013.
\newblock A simple, fast, and effective reparameterization of ibm model 2.
\newblock In \emph{Proceedings of the 2013 Conference of the North American
  Chapter of the Association for Computational Linguistics: Human Language
  Technologies}, pages 644--648.

\bibitem[{Gal and Ghahramani(2016)}]{gal2016theoretically}
Yarin Gal and Zoubin Ghahramani. 2016.
\newblock A theoretically grounded application of dropout in recurrent neural
  networks.
\newblock In \emph{Advances in neural information processing systems}, pages
  1019--1027.

\bibitem[{Gu et~al.(2016)Gu, Lu, Li, and Li}]{gu2016incorporating}
Jiatao Gu, Zhengdong Lu, Hang Li, and Victor~OK Li. 2016.
\newblock Incorporating copying mechanism in sequence-to-sequence learning.
\newblock \emph{arXiv preprint arXiv:1603.06393}.

\bibitem[{Gulcehre et~al.(2016)Gulcehre, Ahn, Nallapati, Zhou, and
  Bengio}]{gulcehre2016pointing}
Caglar Gulcehre, Sungjin Ahn, Ramesh Nallapati, Bowen Zhou, and Yoshua Bengio.
  2016.
\newblock Pointing the unknown words.
\newblock \emph{arXiv preprint arXiv:1603.08148}.

\bibitem[{Hasler et~al.(2018)Hasler, De~Gispert, Iglesias, and
  Byrne}]{hasler2018neural}
Eva Hasler, Adri{\`a} De~Gispert, Gonzalo Iglesias, and Bill Byrne. 2018.
\newblock Neural machine translation decoding with terminology constraints.
\newblock \emph{arXiv preprint arXiv:1805.03750}.

\bibitem[{Hokamp and Liu(2017)}]{hokamp2017lexically}
Chris Hokamp and Qun Liu. 2017.
\newblock Lexically constrained decoding for sequence generation using grid
  beam search.
\newblock \emph{arXiv preprint arXiv:1704.07138}.

\bibitem[{Junczys-Dowmunt et~al.(2016)Junczys-Dowmunt, Dwojak, and
  Sennrich}]{junczys2016amu}
Marcin Junczys-Dowmunt, Tomasz Dwojak, and Rico Sennrich. 2016.
\newblock The amu-uedin submission to the wmt16 news translation task:
  Attention-based nmt models as feature functions in phrase-based smt.
\newblock \emph{arXiv preprint arXiv:1605.04809}.

\bibitem[{Junczys-Dowmunt et~al.(2018)Junczys-Dowmunt, Grundkiewicz, Dwojak,
  Hoang, Heafield, Neckermann, Seide, Germann, Fikri~Aji, Bogoychev, Martins,
  and Birch}]{mariannmt}
Marcin Junczys-Dowmunt, Roman Grundkiewicz, Tomasz Dwojak, Hieu Hoang, Kenneth
  Heafield, Tom Neckermann, Frank Seide, Ulrich Germann, Alham Fikri~Aji,
  Nikolay Bogoychev, Andr\'{e} F.~T. Martins, and Alexandra Birch. 2018.
\newblock \href {http://www.aclweb.org/anthology/P18-4020} {Marian: Fast neural
  machine translation in {C++}}.
\newblock In \emph{Proceedings of ACL 2018, System Demonstrations}, pages
  116--121, Melbourne, Australia. Association for Computational Linguistics.

\bibitem[{Kingma and Ba(2014)}]{kingma2014adam}
Diederik~P Kingma and Jimmy Ba. 2014.
\newblock Adam: A method for stochastic optimization.
\newblock \emph{arXiv preprint arXiv:1412.6980}.

\bibitem[{Koehn et~al.(2007)Koehn, Hoang, Birch, Callison-Burch, Federico,
  Bertoldi, Cowan, Shen, Moran, Zens et~al.}]{koehn2007moses}
Philipp Koehn, Hieu Hoang, Alexandra Birch, Chris Callison-Burch, Marcello
  Federico, Nicola Bertoldi, Brooke Cowan, Wade Shen, Christine Moran, Richard
  Zens, et~al. 2007.
\newblock Moses: Open source toolkit for statistical machine translation.
\newblock In \emph{Proceedings of the 45th annual meeting of the ACL on
  interactive poster and demonstration sessions}, pages 177--180. Association
  for Computational Linguistics.

\bibitem[{Li et~al.(2016)Li, Zhang, and Zong}]{li2016neural}
Xiaoqing Li, Jiajun Zhang, and Chengqing Zong. 2016.
\newblock Neural name translation improves neural machine translation.
\newblock \emph{arXiv preprint arXiv:1607.01856}.

\bibitem[{Luong and Manning(2015)}]{luong2015stanford}
Minh-Thang Luong and Christopher~D Manning. 2015.
\newblock Stanford neural machine translation systems for spoken language
  domains.
\newblock In \emph{Proceedings of the International Workshop on Spoken Language
  Translation}, pages 76--79.

\bibitem[{Luong et~al.(2014)Luong, Sutskever, Le, Vinyals, and
  Zaremba}]{luong2014addressing}
Minh-Thang Luong, Ilya Sutskever, Quoc~V Le, Oriol Vinyals, and Wojciech
  Zaremba. 2014.
\newblock Addressing the rare word problem in neural machine translation.
\newblock \emph{arXiv preprint arXiv:1410.8206}.

\bibitem[{Papineni et~al.(2002)Papineni, Roukos, Ward, and Zhu}]{papineni:2002}
Kishore Papineni, Salim Roukos, Todd Ward, and Wei-Jing Zhu. 2002.
\newblock \href {https://doi.org/10.3115/1073083.1073135} {Bleu: a method for
  automatic evaluation of machine translation}.
\newblock In \emph{Proc. ACL}, pages 311--318, Philadelphia, Pennsylvania, USA.

\bibitem[{Post and Vilar(2018)}]{post2018fast}
Matt Post and David Vilar. 2018.
\newblock Fast lexically constrained decoding with dynamic beam allocation for
  neural machine translation.
\newblock \emph{arXiv preprint arXiv:1804.06609}.

\bibitem[{Ramachandran et~al.(2018)Ramachandran, Zoph, and
  Le}]{ramachandran2018searching}
Prajit Ramachandran, Barret Zoph, and Quoc~V Le. 2018.
\newblock Searching for activation functions.

\bibitem[{See et~al.(2017)See, Liu, and Manning}]{see2017get}
Abigail See, Peter~J Liu, and Christopher~D Manning. 2017.
\newblock Get to the point: Summarization with pointer-generator networks.
\newblock \emph{arXiv preprint arXiv:1704.04368}.

\bibitem[{Sennrich et~al.(2015{\natexlab{a}})Sennrich, Haddow, and
  Birch}]{Sennrich2015Improving}
Rico Sennrich, Barry Haddow, and Alexandra Birch. 2015{\natexlab{a}}.
\newblock Improving neural machine translation models with monolingual data.
\newblock \emph{Computer Science}.

\bibitem[{Sennrich et~al.(2015{\natexlab{b}})Sennrich, Haddow, and
  Birch}]{sennrich2015neural}
Rico Sennrich, Barry Haddow, and Alexandra Birch. 2015{\natexlab{b}}.
\newblock Neural machine translation of rare words with subword units.
\newblock \emph{arXiv preprint arXiv:1508.07909}.

\bibitem[{Tang et~al.(2016)Tang, Meng, Lu, Li, and Yu}]{tang2016neural}
Yaohua Tang, Fandong Meng, Zhengdong Lu, Hang Li, and Philip~LH Yu. 2016.
\newblock Neural machine translation with external phrase memory.
\newblock \emph{arXiv preprint arXiv:1606.01792}.

\bibitem[{Vaswani et~al.(2017)Vaswani, Shazeer, Parmar, Uszkoreit, Jones,
  Gomez, Kaiser, and Polosukhin}]{DBLP:journals/corr/VaswaniSPUJGKP17}
Ashish Vaswani, Noam Shazeer, Niki Parmar, Jakob Uszkoreit, Llion Jones,
  Aidan~N. Gomez, Lukasz Kaiser, and Illia Polosukhin. 2017.
\newblock \href {http://arxiv.org/abs/1706.03762} {Attention is all you need}.
\newblock \emph{CoRR}, abs/1706.03762.

\bibitem[{Vinyals et~al.(2015)Vinyals, Fortunato, and
  Jaitly}]{vinyals2015pointer}
Oriol Vinyals, Meire Fortunato, and Navdeep Jaitly. 2015.
\newblock Pointer networks.
\newblock In \emph{Advances in Neural Information Processing Systems}, pages
  2692--2700.

\bibitem[{Wang et~al.(2017{\natexlab{a}})Wang, Lu, Tu, Li, Xiong, and
  Zhang}]{wang2017neural}
Xing Wang, Zhengdong Lu, Zhaopeng Tu, Hang Li, Deyi Xiong, and Min Zhang.
  2017{\natexlab{a}}.
\newblock Neural machine translation advised by statistical machine
  translation.
\newblock In \emph{AAAI}, pages 3330--3336.

\bibitem[{Wang et~al.(2017{\natexlab{b}})Wang, Cheng, Jiang, Yang, Chen, Li,
  Shi, Wang, and Yang}]{wang2017sogou}
Yuguang Wang, Shanbo Cheng, Liyang Jiang, Jiajun Yang, Wei Chen, Muze Li, Lin
  Shi, Yanfeng Wang, and Hongtao Yang. 2017{\natexlab{b}}.
\newblock Sogou neural machine translation systems for wmt17.
\newblock In \emph{Proceedings of the Second Conference on Machine
  Translation}, pages 410--415.

\bibitem[{Zens et~al.(2012)Zens, Stanton, and Xu}]{zens2012systematic}
Richard Zens, Daisy Stanton, and Peng Xu. 2012.
\newblock A systematic comparison of phrase table pruning techniques.
\newblock In \emph{Proceedings of the 2012 Joint Conference on Empirical
  Methods in Natural Language Processing and Computational Natural Language
  Learning}, pages 972--983. Association for Computational Linguistics.

\end{thebibliography}

\end{document}